\def\year{2018}\relax
\documentclass[letterpaper]{article} 
\usepackage{aaai18}  
\usepackage{times}  
\usepackage{helvet}  
\usepackage{courier}  
\usepackage{url}  
\usepackage{graphicx}  
\usepackage{subcaption}
\usepackage{amsmath}
\usepackage{amssymb}
\usepackage{color}
\usepackage{multirow}
\newcommand\energy{\mathcal{E}}  
\newcommand\best[1]{{\textbf{#1}}}

\makeatletter
\newcommand{\thickhline}{%
    \noalign {\ifnum 0=`}\fi \hrule height 1pt
    \futurelet \reserved@a \@xhline
}
\makeatother

\begin{document}

\title{Scene-centric Joint Parsing of Cross-view Videos}
\author{Hang Qi$^{1}$\thanks{Hang Qi, Yuanlu Xu and Tao Yuan contributed equally to this paper. This work is supported by ONR MURI project N00014-16-1-2007, DARPA XAI Award N66001-17-2-4029, and NSF IIS 1423305.},\, Yuanlu Xu$^{1*}$, \, Tao Yuan$^{1*}$, Tianfu Wu$^{2}$, \, Song-Chun Zhu$^1$\\
$^1$Dept. Computer Science and Statistics, University of California, Los Angeles (UCLA)\\
$^2$Dept. Electrical and Computer Engineering, NC State University\\
{\tt\small \{hangqi, yuanluxu\}@cs.ucla.edu, taoyuan@ucla.edu, tianfu\_wu@ncsu.edu, sczhu@stat.ucla.edu}
}

\maketitle

\begin{abstract}

Cross-view video understanding is an important yet under-explored area in computer vision. In this paper, we introduce a joint parsing framework that integrates view-centric proposals into scene-centric parse graphs that represent a coherent scene-centric understanding of cross-view scenes. Our key observations are that overlapping fields of views embed rich appearance and geometry correlations and that knowledge fragments corresponding to individual vision tasks are governed by consistency constraints available in commonsense knowledge. The proposed joint parsing framework represents such correlations and constraints explicitly and generates semantic scene-centric parse graphs. Quantitative experiments show that scene-centric predictions in the parse graph outperform view-centric predictions.

\end{abstract}

\section{Introduction}

During the past decades, remarkable progress has been made in many vision tasks, e.g., image classification, object detection, pose estimation. Recently, more comprehensive visual tasks probe deeper understanding of visual scenes under interactive and multi-modality settings, such as visual Turing tests~\cite{geman2015,qi2015restricted} and visual question answering~\cite{vqa2015}. In addition to discriminative tasks focusing on binary or categorical predictions, emerging research involves representing fine-grained relationships in visual scenes~\cite{krishnavisualgenome,yangdeepiu} and unfolding semantic structures in contexts including caption or description generation~\cite{yao2010i2t}, and question answering~\cite{Tu13,zhu2016cvpr}.

\begin{figure}[ptb]
\centering
\includegraphics[width=\linewidth]{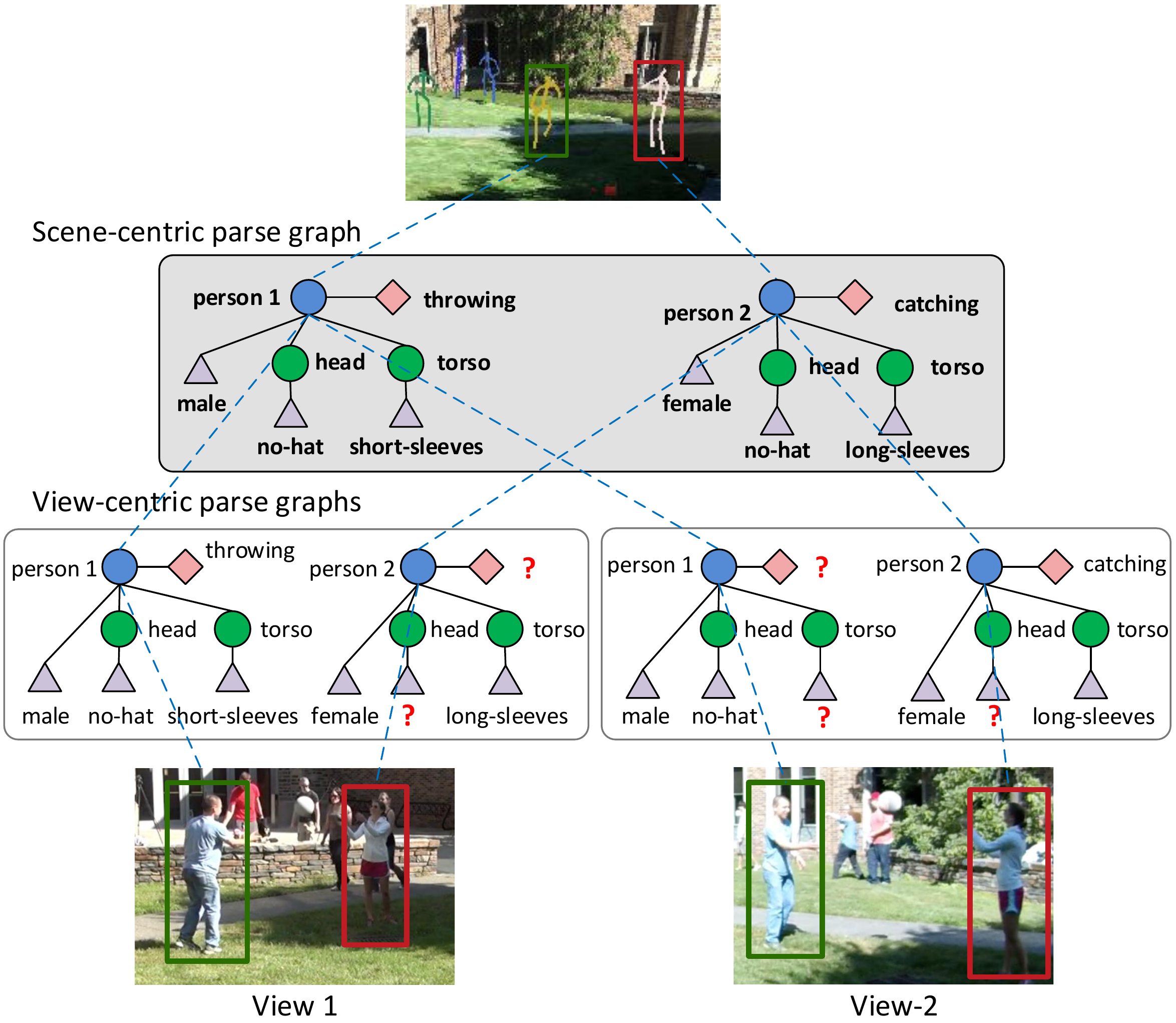}
\caption{An example of the spatio-temporal semantic parse graph hierarchy in a visual scene captured by two cameras.}
\label{fig:intro}
\end{figure}

In this paper, we present a framework for uncovering the semantic structure of scenes in a cross-view camera network. The central requirement is to resolve ambiguity and establish cross-reference among information from multiple cameras. Unlike images and videos shot from single static point of view, cross-view settings embed rich physical and geometry constraints due to the overlap between fields of views.
While multi-camera setups are common in real-word surveillance systems, large-scale cross-view activity dataset are not available due to privacy and security reasons. This makes data-demanding deep learning approaches infeasible.

Our joint parsing framework computes a hierarchy of spatio-temporal parse graphs by establishing cross-reference of entities among different views and inferring their semantic attributes from a scene-centric perspective. For example, Fig.~\ref{fig:intro} shows a parse graph hierarchy that describes a scene where two people are playing a ball. In the first view, person 2's action is not grounded because of the cluttered background, while it is detected in the second view. Each view-centric parse graph contains local recognition decisions in an individual view, and the scene centric parse graph summaries a comprehensive understanding of the scene with coherent knowledge.

The structure of each individual parse graph fragment is induced by an ontology graph that regulates the domain of interests. A parse graph hierarchy is used to represent the correspondence of entities between the multiple views and the scene. We use a probabilistic model to incorporate various constraints on the parse graph hierarchy and formulate the joint parsing as a MAP inference problem. A MCMC sampling algorithm and a dynamic programming algorithm are used to explore the joint space of scene-centric and view-centric interpretations and to optimize for the optimal solutions. Quantitative experiments show that scene-centric parse graphs outperforms the initial view-centric proposals.

{\bf Contributions}. The contributions of this work are three-fold: (i) a unified hierarchical parse graph representation for cross-view person, action, and attributes recognition;
(ii) a stochastic inference algorithm that explores the joint space of scene-centric and view-centric interpretations efficiently starting with initial proposals;
(iii) a joint parse graph hierarchy that is an interpretable representation for scene and events.

\section{Related Work} \label{sec:literature}

Our work is closely related to three research areas in computer vision and artificial intelligence.

\textbf{Multi-view video analytics.} Typical multi-view visual analytics tasks include object detection~\cite{mvdetectCVPR10,UtasiCVPR2011}, cross-view tracking~\cite{BerclazTPAMI2011,Leal-TaixeCVPR2012,xu2016multi,xu2017multi}, action recognition~\cite{wangcross}, person re-identification~\cite{xu2013reid,xu2014search} and 3D reconstruction~\cite{HofmannCVPR2013}.
While heuristics such as appearances and motion consistency constraints have been used to regularize the solution space, these methods focus on a specific multi-view vision task whereas we aim to propose a general framework to jointly resolve a wide variety of tasks.

\textbf{Semantic representations.} Semantic and expressive representations have been developed for various vision tasks, e.g., image parsing~\cite{HanTPAMI2009}, 3D scene reconstruction~\cite{liusingle,PeroCVPR13}, human-object interaction~\cite{koppula2016anticipating}, pose and attribute estimation~\cite{Ping4DHOI16}. In this paper, our representation also falls into this category. The difference is that our model is defined upon cross-view spatio-temporal domain and is able to incorporate a variety of tasks.

\textbf{Interpretability.} Automated generation of explanations regarding predictions has a long and rich history in artificial intelligence. Explanation systems have been developed for a wide range of applications, including simulator actions~\cite{van2004explainable,lane2005explainable,core2006building}, robot movements~\cite{lomas2012explaining}, and object recognition in images~\cite{biran2014justification,hendricks2016generating}. Most of these approaches are rule-based and suffer from generalization across different domains. Recent methods including~\cite{ribeiro2016should} use proxy models or data to interpret black box models, while our scene-centric parse graphs are explicit representations of the knowledge by definition.

\section{Representation}  \label{sec:repre}

\begin{figure}[ptb]
\centering
\includegraphics[width=\linewidth]{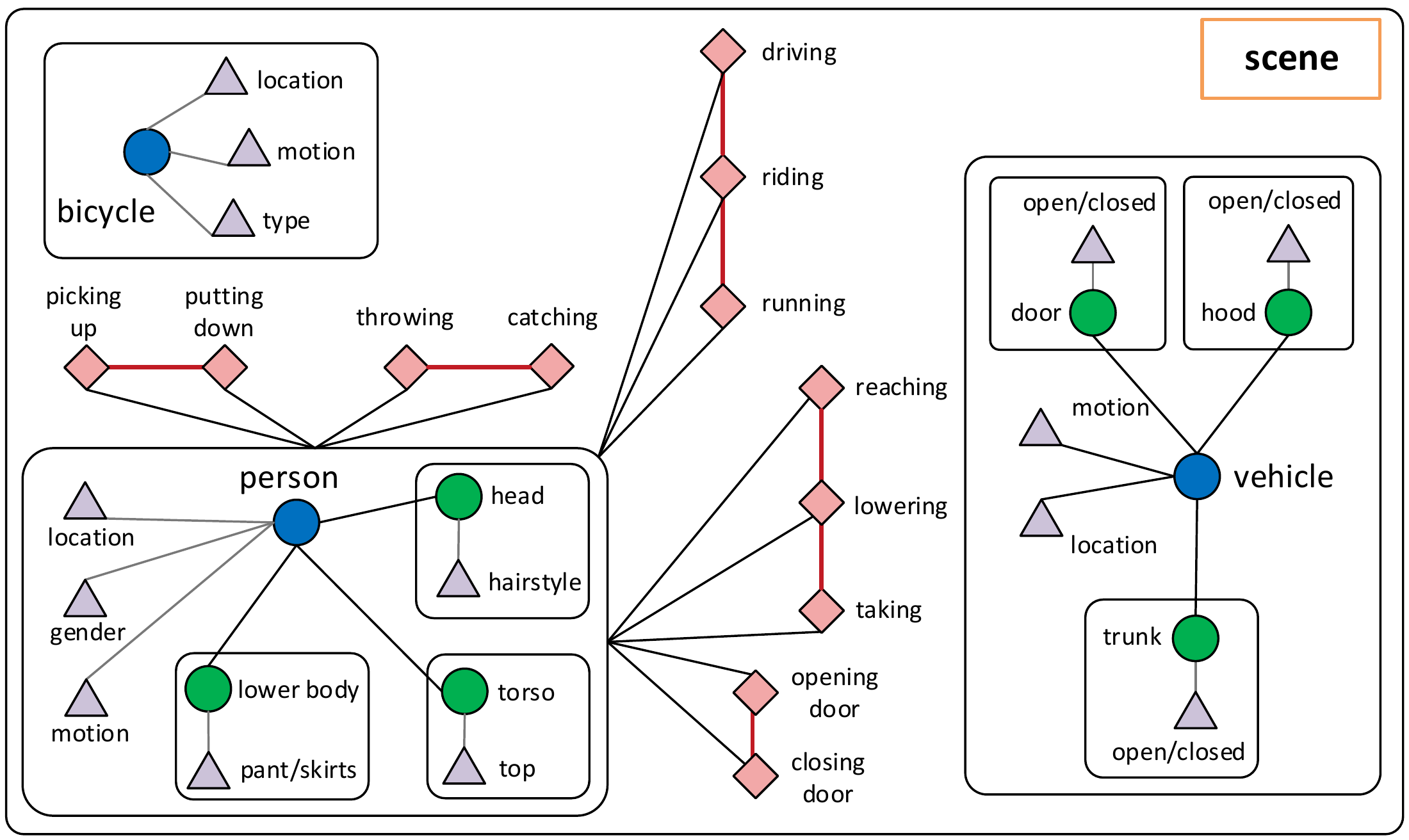}
\caption{An illustration of the proposed ontology graph describing objects, parts, actions and attributes.}
\label{fig:ontology}
\end{figure}

A scene-centric spatio-temporal parse graph represents humans, their actions and attributes, interaction with other objects captured by a network of cameras.
We will first introduce the concept of ontology graph as domain definitions, then we will describe parse graphs and parse graph hierarchy as view-centric and scene-centric representations respectively.

\begin{figure*}[ptb]
\centering
\includegraphics[width=0.825\linewidth]{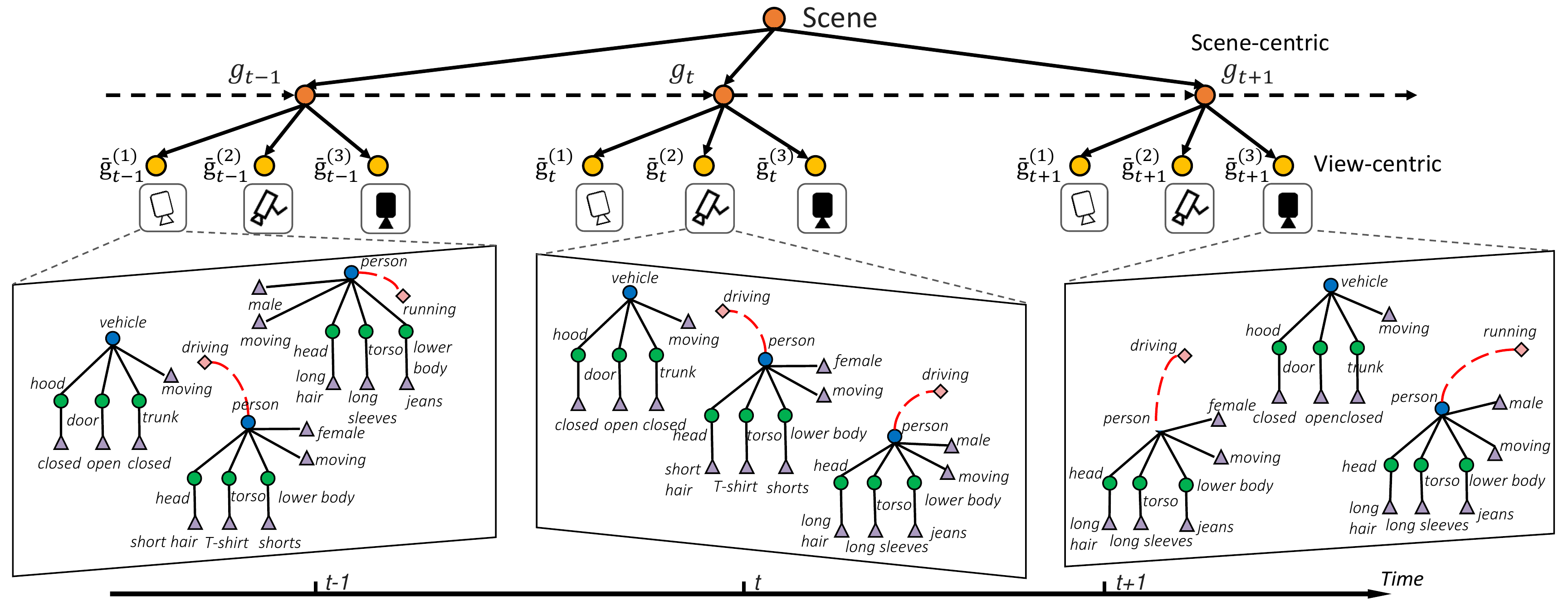}
\caption{The proposed spatio-temporal parse graph hierarchy. (Better viewed electronically and zoomed).}
\label{fig:dag}
\end{figure*}

\textbf{Ontology graph}. To define the scope of our representation on scenes and events, an ontology is used to describe a set of plausible objects, actions and attributes.
We define an ontology as a graph that contains nodes representing objects, parts, actions, attributes respectively and edges representing the relationships between nodes. Specifically, every object and part node is a concrete type of object that can be detected in videos.
Edges between object and part nodes encodes ``part-of'' relationships.
Action and attribute nodes connected to an object or part node represent plausible actions and appearance attributes the object can take. For example, Fig.~\ref{fig:ontology} shows an ontology graph that describes a domain including people, vehicles, bicycles. An object can be decomposed into parts (i.e., green nodes), and enriched with actions (i.e., pink nodes) and attributes (i.e., purple diamonds). The red edges among action nodes denote their incompatibility. The ontology graph can be considered a compact AOG~\cite{liusingle,Ping4DHOI16} without the compositional relationships and event hierarchy. In this paper, we focus on a restricted domain inspired by~\cite{qi2015restricted}, while larger ontology graphs can be easily derived from large-scale visual relationship datasets such as~\cite{krishnavisualgenome} and open-domain knowledge bases such as~\cite{liu2004conceptnet}.

\textbf{Parse graphs}. While an ontology describes plausible elements, only a subset of these concepts can be true for a given instance at a given time. For example, a person cannot be both ``standing'' and ``sitting'' at the same time, while both are plausible actions that a person can take. To distinguish plausible facts and satisfied facts, we say a node is \textit{grounded} when it is associated with data. Therefore, a subgraph of the ontology graph that only contains grounded nodes can be used to represent a specific \textit{instance} (e.g. a specific person) at a specific time. In this paper, we refer to such subgraphs as \textit{parse graphs}.

\textbf{Parse graph hierarchy}. In cross-view setups, since each view only captures an incomplete set of facts in a scene, we use a spatio-temporal hierarchy of parse graphs to represent the collective knowledge of the scene and all the individual views. To be concrete, a view-centric parse graph $\tilde{g}$ contains nodes grounded to a video sequence captured by an individual camera, whereas a scene-centric parse graph $g$ is an aggregation of view-centric parse graphs and therefore reflects a global understanding of the scene. As illustrated in Fig.~\ref{fig:dag}, for each time step $t$, the scene-centric parse graph $g_t$ is connected with the corresponding view-centric parse graphs $\tilde{g}_t^{(i)}$ indexed by the views, and the scene-centric graphs are regarded as a Markov chain in the temporal sequence. In terms of notations, in this paper we use a tilde notation to represent the view-centric concepts $\tilde{x}$ corresponding to scene-centric concepts $x$.

\section{Probabilistic Formulation}  \label{sec:formulation}

The task of joint parsing is to infer the spatio-temporal parse graph hierarchy $G = \langle \Phi, g, \tilde{g}^{(1)}, \tilde{g}^{(2)}, \ldots, \tilde{g}^{(M)}\rangle $ from the input frames from video sequences $I = \{I_t^{(i)}\}$ captured by a network of $M$ cameras
, where $\Phi$ is an object identity mapping between scene-centric parse graph $g$ and view-centric parse graphs $\tilde{g}^{(i)}$ from camera $i$. $\Phi$ defines the structure of parse graph hierarchy. In this section, we discuss the formulation assuming a fixed structure, while defer the discussion of how to traverse the solution space to section~\ref{sec:inference}.

We formulate the inference of parse graph hierarchy as a MAP inference problem in a posterior distribution $p( G | I)$ as follows
\begin{equation}\begin{aligned}\label{eqn:posterior}
G^* &= \arg\max_{G} p(I | G) \cdot p(G).
\end{aligned} \end{equation}

\textbf{Likelihood.} The likelihood term models the grounding of nodes in view-centric parse graphs to the input video sequences. Specifically,
\begin{equation} \begin{aligned}
p(I | G) &= \prod_{i=1}^{M} \prod_{t=1}^{T} p(I_t^{(i)} | \tilde{g}_{t}^{(i)}) \\
&= \prod_{i=1}^{M} \prod_{t=1}^{T} \prod_{v \in V(\tilde{g}_{i}^{(t)})} p(I(v) | v),
\label{eq:likelihood}
\end{aligned} \end{equation}
where $\tilde{g}_t^{(i)}$ is the view-centric parse graph of camera $i$ at time $t$ and
$V(\tilde{g}_{t}^{(i)})$ is the set of nodes in the parse graph. $p(I(v) | v)$ is the node likelihood for the concept represented by node $v$ being grounded on the data fragment $I(v)$. In practice, this probability can be approximated by normalized detection and classifications scores~\cite{PirsiavashCVPR2011}.

\textbf{Prior.} The prior term models the compatibility of scene-centric and view-centric parse graphs across time. We factorize the prior as
\begin{align}
p(G) = & p(g_1)\prod_{t=1}^{T-1} p(g_{t+1} | g_{t}) \,\prod_{i=1}^{M} \prod_{t=1}^{T} p(\tilde{g}_t^{(i)} | g_t),
\end{align}
where $p(g_1)$ is a prior distribution on parse graphs that regulates the combination of nodes, and $p(g_t | g_{t-1})$ is a transitions probability of scene-centric parse graphs across time. Both probability distributions are estimated from training sequences. $p(\tilde{g}_t^{(i)} | g_t)$ is defined as a Gibbs distribution that models the compatibility of scene-centric and view-centric parse graphs in the hierarchy (we drop subscripts $t$ and camera index $i$ for brevity).
\begin{equation} \begin{aligned}
p(\tilde{g} | g) &= \dfrac{1}{Z} \exp \{ - \energy(g, \tilde{g}) \} \\
	&= \dfrac{1}{Z} \exp \{ - w_1\energy_S(g, \tilde{g}) - w_2\energy_A(g, \tilde{g}) \\
	& \quad\quad\quad\quad\; - w_3\energy_{Act}(g, \tilde{g}) - w_4\energy_{Attr}(g, \tilde{g}) \},
\end{aligned} \end{equation}
where energy $\energy(g, \tilde{g})$ is decomposed into four different terms described in detail in the subsection below. The weights are tuning parameters that can be learned via cross-validation. We consider view-centric parse graphs for videos from different cameras are independent conditioned on scene-centric parse graph under the assumption that all cameras have fixed and known locations.

\subsection{Cross-view Compatibility}

In this subsection, we describe the energy function $\energy(g, \tilde{g})$ for regulating the compatibility between the occurrence of objects in the scene and an individual view from various aspects. Note that we use a tilde notation to represent the node correspondence in scene-centric and view-centric parse graphs (i.e., for a node $v\in g$ in a scene-centric parse graph, we refer to the corresponding node in a view-centric parse graph as $\tilde{v}$).

\textbf{Appearance similarity.} For each object node in the parse graph, we keep an appearance descriptor. The appearance energy regulates the appearance similarity of object $o$ in the scene-centric parse graph and $\tilde{o}$ in the view-centric parse graphs.
\begin{align}
\energy_A(g, \tilde{g}) = \sum_{o \in g} || (\phi(o) - \phi(\tilde{o}) ||_2,
\end{align}
where $\phi(\cdot)$ is the appearance feature vector of the object. At the view-level, this feature vector can be extracted by pre-trained convolutional neural networks; at the scene level, we use a mean pooling of view-centric features.

\textbf{Spatial consistency.} At each time point, every object in a scene has a fixed physical location in the world coordinate system while appears on the image plane of each camera according to the camera projection. For each object node in the parse graph hierarchy, we keep a scene-centric location $s(o)$ for each object $o$ in scene-centric parse graphs and a view-centric location $s(\tilde{o})$ on the image plane in view-centric parse graphs. The following energy is defined to enforce the spatial consistency:
\begin{align}
\energy_S(g, \tilde{g}) = \sum_{o \in g} ||s(o) - h(s(\tilde{o}))||_2,
\end{align}
where $h(\cdot)$ is a perspective transform that maps a person's view-centric foot point coordinates to the world coordinates on the ground plane of the scene with the camera homography, which can be obtained via the intrinsic and extrinsic camera parameters.

\textbf{Action compatibility.} Among action and object part nodes, scene-centric human action predictions shall agree with the human pose observed in individual views from different viewing angles:
\begin{align}
\energy_{Act}(g, \tilde{g}) = \sum_{l \in g}-\log p( l | \tilde{p}),
\end{align}
where $l$ is an action node in scene-centric parse graphs and $\tilde{p}$ are positions of all human parts in the view-centric parse graph. In practice, we separately train a action classifier that predicts action classes with joint positions of human parts and uses the classification score to approximate this probability.

\textbf{Attribute consistency.} In cross-view sequences, entities observed from multiple cameras shall have a consistent set of attributes. This energy term models the commonsense constraint that scene-centric human attributes shall agree with the observation in individual views:
\begin{align}
\energy_{Attr}(g, \tilde{g}) = \sum_{a \in g} {\bf 1}(a\neq\tilde{a})\cdot \xi,
\end{align}
where ${\bf 1}(\cdot)$ is an indicator function and $\xi$ is a constant energy penalty introduced when the two predictions mismatch.

\section{Inference}  \label{sec:inference}

The inference process consists of two sub-steps: (i) matching object nodes $\Phi$ in scene-centric and view-centric parse graphs (i.e. the structure of parse graph hierarchy) and (ii) estimating optimal values of parse graphs $\{g, \tilde{g}^{(1)},\dots,\tilde{g}^{(M)}\}$.

The overall procedure is as follows: we first obtain view-centric objects, actions, and attributes proposals from pre-trained detectors on all video frames. This forms the initial view-centric predictions $\{\tilde{g}^{(1)},\dots,\tilde{g}^{(M)}\}$. Next we use a Markov Chain Monte Carlo (MCMC) sampling algorithm to optimize the parse graph structure $\Phi$. Given a fixed parse graph hierarchy, variables within the scene-centric and view-centric parse graphs $\{g, \tilde{g}^{(1)},\dots,\tilde{g}^{(M)}\}$ can be efficiently estimated by a dynamic programming algorithm. These two steps are performed iteratively until convergence.

\subsection{Inferring Parse Graph Hierarchy}

We use a stochastic algorithm to traverse the solution space of the parse graph hierarchy $\Phi$. To satisfy the detailed balance condition, we define three reversible operators $\Theta = \{\Theta_1, \Theta_2, \Theta_3\}$ as follows.

\textbf{Merging}. The merging operator $\Theta_1$ groups a view-centric parse graph with an other view-centric parse graph by creating a scene-centric parse graph that connects the two. The operator requires the two operands to describe two objects of the same type either from different views or in the same view but with non-overlapping time intervals.

\textbf{Splitting}. The splitting operator $\Theta_2$ splits a scene-centric parse graph into two parse graphs such that each resulting parse graph only connects to a subset of view-centric parse graphs.

\textbf{Swapping}. The swapping operator $\Theta_3$ swaps two view-centric parse graphs. One can view the swapping operator as a shortcut of merging and splitting combined.

We define the proposal distribution $q(G \rightarrow G')$ as an uniform distribution. At each iteration, we generate a new structure proposal $\Phi'$ by applying one of the three operators $\Theta_i$ with respect to probability 0.4, 0.4, and 0.2, respectively. The generated proposal is then accepted with respect to an acceptance rate $\alpha(\cdot)$ as in the Metropolis-Hastings algorithm~\cite{MHAlgorithm}:
\begin{equation}
\alpha(G \rightarrow G') = \min\left(1,\, \dfrac{q(G' \rightarrow G) \cdot p(G'|x)}{q(G \rightarrow G') \cdot p(G|x)}\right),
\end{equation}
where $p(G|x)$ the posterior is defined in Eqn. (\ref{eqn:posterior}).

\subsection{Inferring Parse Graph Variables}

Given a fixed parse graph hierarchy, we need to estimate the optimal value for each node within each parse graph.
As illustrated in Fig.~\ref{fig:dag}, for each frame, the scene-centric node $g_t$ and the corresponding view-centric nodes $\tilde{g}_t^{(i)}$ form a star model, and the whole scene-centric nodes are regarded as a Markov chain in the temporal order. Therefore the proposed model is essentially a Directed Acyclic Graph (DAG). To infer the optimal node values, we can simply apply the standard factor graph belief propagation (sum-product) algorithm.

\section{Experiments}

\subsection{Setup and Datasets}

We evaluate our scene-centric joint-parsing framework in tasks including object detection, multi-object tracking, action recognition, and human attributes recognition. In object detection and multi-object tracking tasks, we compare with published results. In action recognition and human attributes tasks, we compare the performance of view-centric proposals without joint parsing and scene-centric predictions after joint parsing as well as additional baselines. The following datasets are used to cover a variety of tasks.

The \textbf{CAMPUS} dataset~\cite{xu2016multi}
\footnote{bitbucket.org/merayxu/multiview-object-tracking-dataset} contains video sequences from four scenes each captured by four cameras. Different from other multi-view video datasets focusing solely on multi-object tracking task, videos in the CAMPUS dataset contains richer human poses and activities with moderate overlap in the fields of views between cameras. In addition to the tracking annotation in the CAMPUS dataset, we collect new annotation that includes 5 action categories and 9 attribute categories for evaluating action and attribute recognition.

The \textbf{TUM Kitchen} dataset~\cite{tenorth2009tum}\footnote{ias.in.tum.de/software/kitchen-activity-data}
is an action recognition dataset that contains 20 video sequences captured by 4 cameras with overlapping views. As we only focusing on the RGB imagery inputs in our framework, other modalities such as motion capturing, RFID tag reader signals, magnetic sensor signals are not used as inputs in our experiments. To evaluate detection and tracking task, we compute human bounding boxes from motion capturing data by projecting 3D human poses to the image planes of all cameras using the intrinsic and extrinsic parameters provided in the dataset. To evaluate human attribute tasks, we annotate 9 human attribute categories for every subject.

In our experiments, both the CAMPUS and the TUM Kitchen datasets are used in all tasks.
In the following subsection, we present isolated evaluations.

\subsection{Evaluation}


\begin{table}[ptb]
\begin{center}
\resizebox{\linewidth}{!}{
\begin{tabular}{l|cccccc}
\hline\thickhline
CAMPUS-S1  & DA (\%) & DP (\%) & TA (\%) & TP (\%) & IDSW & FRAG \\
\hline
Fleuret et al. & 24.52 & 64.28 & 22.43 & 64.17 & 2269 & 2233 \\
Berclaz et al. & 30.47 & 62.13 & 28.10 & 62.01 & 2577 & 2553 \\
Xu et al.      & 49.30 & 72.02 & 56.15 & 72.97 & 320  & 141  \\    		
Ours           & \best{56.00} & 72.98 & \best{55.95} & 72.77 & 310  & 138  \\   		
\thickhline
CAMPUS-S2  & DA (\%) & DP (\%) & TA (\%) & TP (\%) & IDSW & FRAG \\
\hline
Fleuret et al. & 16.51 & 63.92 & 13.95 & 63.81 & 241  & 214 \\
Berclaz et al. & 24.35 & 61.79 & 21.87 & 61.64 & 268  & 249 \\
Xu et al.      & 27.81 & 71.74 & 28.74 & 71.59 & 1563 & 443 \\    		
Ours           & \best{28.24} & 71.49 & \best{27.91} & 71.16 & 1615 & 418 \\   		
\thickhline
CAMPUS-S3  & DA (\%) & DP (\%) & TA (\%) & TP (\%) & IDSW & FRAG \\
\hline
Fleuret et al. & 17.90 & 61.19 & 16.15 & 61.02 & 249 & 235 \\
Berclaz et al. & 19.46 & 59.45 & 17.63 & 59.29 & 264 & 257 \\
Xu et al.      & 49.71 & 67.02 & 49.68 & 66.98 & 219 & 117 \\    		
Ours           & \best{50.60} & 67.00 & \best{50.55} & 66.96 & 212 & 113 \\    		
\thickhline
CAMPUS-S4  & DA (\%) & DP (\%) & TA (\%) & TP (\%) & IDSW & FRAG \\
\hline
Fleuret et al. & 11.68 & 60.10 & 11.00 & 59.98 & 828 & 812 \\
Berclaz et al. & 14.73 & 58.51 & 13.99 & 58.36 & 893 & 880 \\
Xu et al.      & 24.46 & 66.41 & 24.08 & 68.44 & 962 & 200 \\         	
Ours           & \best{24.81} & 66.59 & \best{24.63} & 68.28 & 938 & 194 \\    		
\thickhline
TUM Kitchen   & DA (\%) & DP (\%) & TA (\%) & TP (\%) & IDSW & FRAG \\
\hline
Fleuret et al. & 69.88 & 64.54  & 69.67 & 64.76 & 61 & 57 \\
Berclaz et al. & 72.39 & 63.27  & 72.20 & 63.51 & 48 & 44 \\
Xu et al.      & 86.53 & 72.12  & 86.18 & 72.37 & 9  & 5 \\
Ours           & \best{89.13} & 72.21  & \best{88.77} & 72.42 & 12 & 8 \\
\thickhline
\end{tabular}}
\end{center}
\caption{Quantitative comparisons of multi-object tracking on CAMPUS and TUM Kitchen datasets.}
\label{tab:tracking}
\end{table}

\textbf{Object detection \& tracking}. We use FasterRCNN~\cite{fasterRCNN} to create initial object proposals on all video frames. The detection scores are used in the likelihood term in Eqn. (\ref{eq:likelihood}). During joint parsing, objects which are not initially detected on certain views are projected from object's scene-centric positions with the camera matrices. After joint parsing, we extract all bounding boxes that are grounded by object nodes from each view-centric parse graph to compute multi-object detection accuracy (DA) and precision (DP). Concretely, the accuracy measures the faction of correctly detected objects among all ground-truth objects and the precision is computed as fraction of true-positive predictions among all output predictions. A predicted bounding box is considered a match with a ground-truth box only if the intersection over union (IoU) score is greater than 0.5. When more than one prediction overlaps with a ground-truth box, only the one with the maximum overlap is counted as true positive.

When extracting all bounding boxes on which the view-centric parse graphs are grounded and grouping them according to the identity correspondence between different views, we obtain object trajectories with identity matches across multiple videos. In the evaluation, we compute four major tracking metrics: multi-object tracking accuracy (TA), multi-object track precision (TP), the number of identity switches (IDSW), and the number of fragments (FRAG).
A higher value of TA and TP and a lower value of IDSW and FRAG indicate the tracking method works better. We report quantitative comparisons with several published methods~\cite{xu2016multi,BerclazTPAMI2011,FleuretTPAMI2008} in Table~\ref{tab:tracking}. From the results, the performance measured by tracking metrics are comparable to published results. We conjecture that the appearance similarity is the main drive for establish cross-view correspondence while additional semantic attributes proved limited gain to the tracking task.

\begin{table*}[ptb]
\begin{center}
\setlength\tabcolsep{2pt}
\resizebox{\linewidth}{!}{
\begin{tabular}{l|cccccc||ccccccccc}
\thickhline
\multicolumn{1}{c|}{}  & \multicolumn{6}{c||}{\textbf{CAMPUS}} & \multicolumn{9}{c}{\textbf{TUM Kitchen}} \\
Methods  & Run & PickUp & PutDown & Throw & Catch & Overall
 & Reach & Taking & Lower & Release & OpenDoor &  CloseDoor &  OpenDrawer & CloseDrawer & Overall \\
\hline\cline{1-16}
view-centric &  0.83 &  0.76 &  0.91 &  0.86 &  0.80 &  0.82   & 0.78 &  0.66 &  0.75 &  0.67 &  0.48 &  0.50 &  0.50 &  0.42 &  0.59  \\
baseline-vote & 0.85 &   0.80 &  0.71 &  0.88 &  0.82 & 0.73   & 0.80 &  0.63 &  0.77 &  0.71 &  0.72 &  0.73 &  0.70 &  0.47 &  0.69  \\
baseline-mean & 0.86 &  0.82 &  1.00 &  0.90 &  0.87 &  0.88   & 0.79 &  0.61 &  0.75 &  0.69 &  0.67 &  0.67 &  0.66 &  0.45 &  0.66  \\
scene-centric & 0.87 &  0.83 &  1.00 &  0.91 &  0.88 & \best{0.90}    & 0.81 &  0.67 &  0.79 &  0.71 &  0.71 &  0.73 &  0.70 &  0.50 &  \best{0.70}  \\
\thickhline
\end{tabular}
}
\end{center}
\caption{Quantitative comparisons of human action recognition on CAMPUS and TUM Kitchen datasets.}
\label{tab:action}
\end{table*}

\begin{figure}[ptb]
\centering
\includegraphics[width=0.495\linewidth]{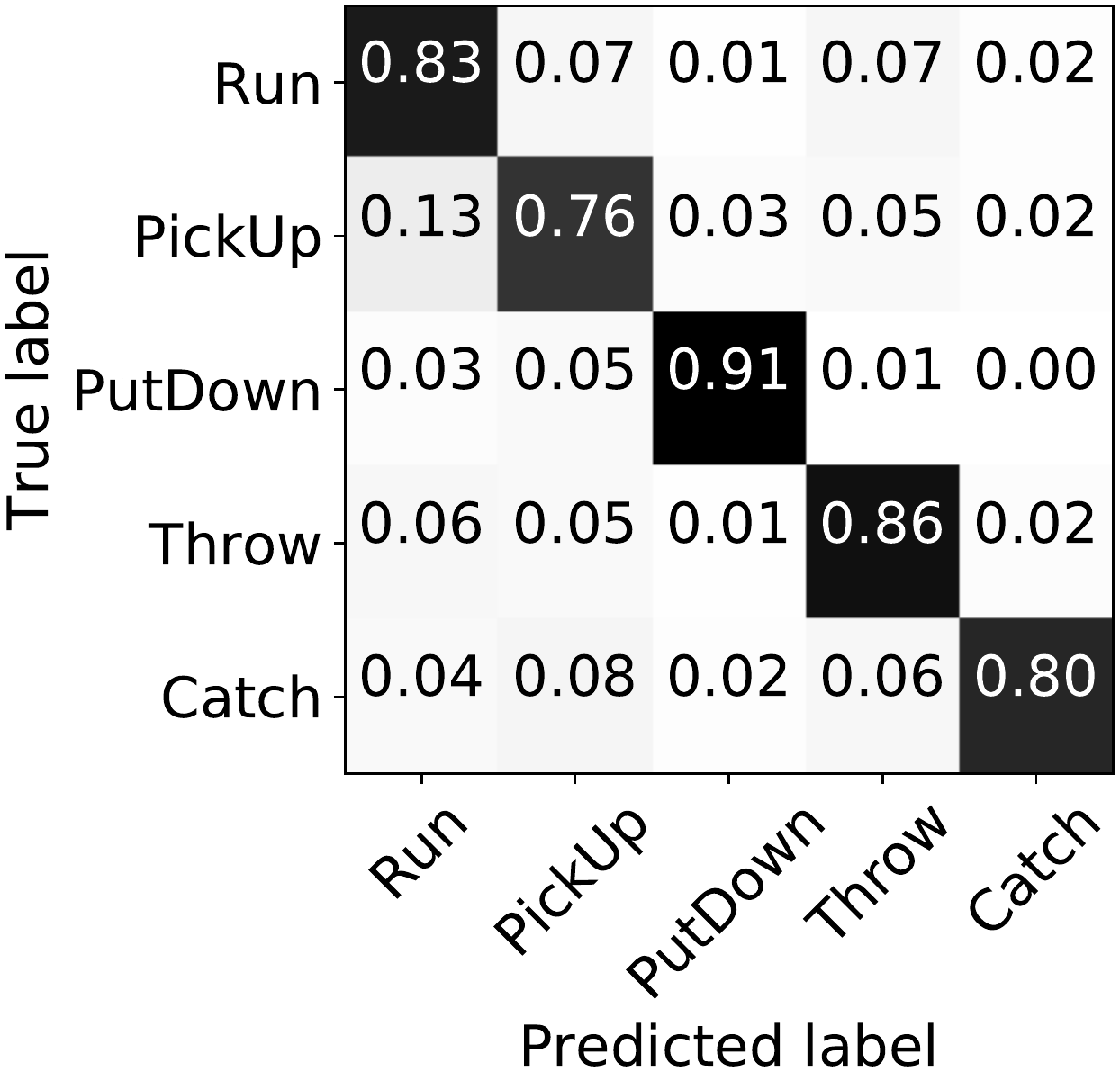}
\includegraphics[width=0.495\linewidth]{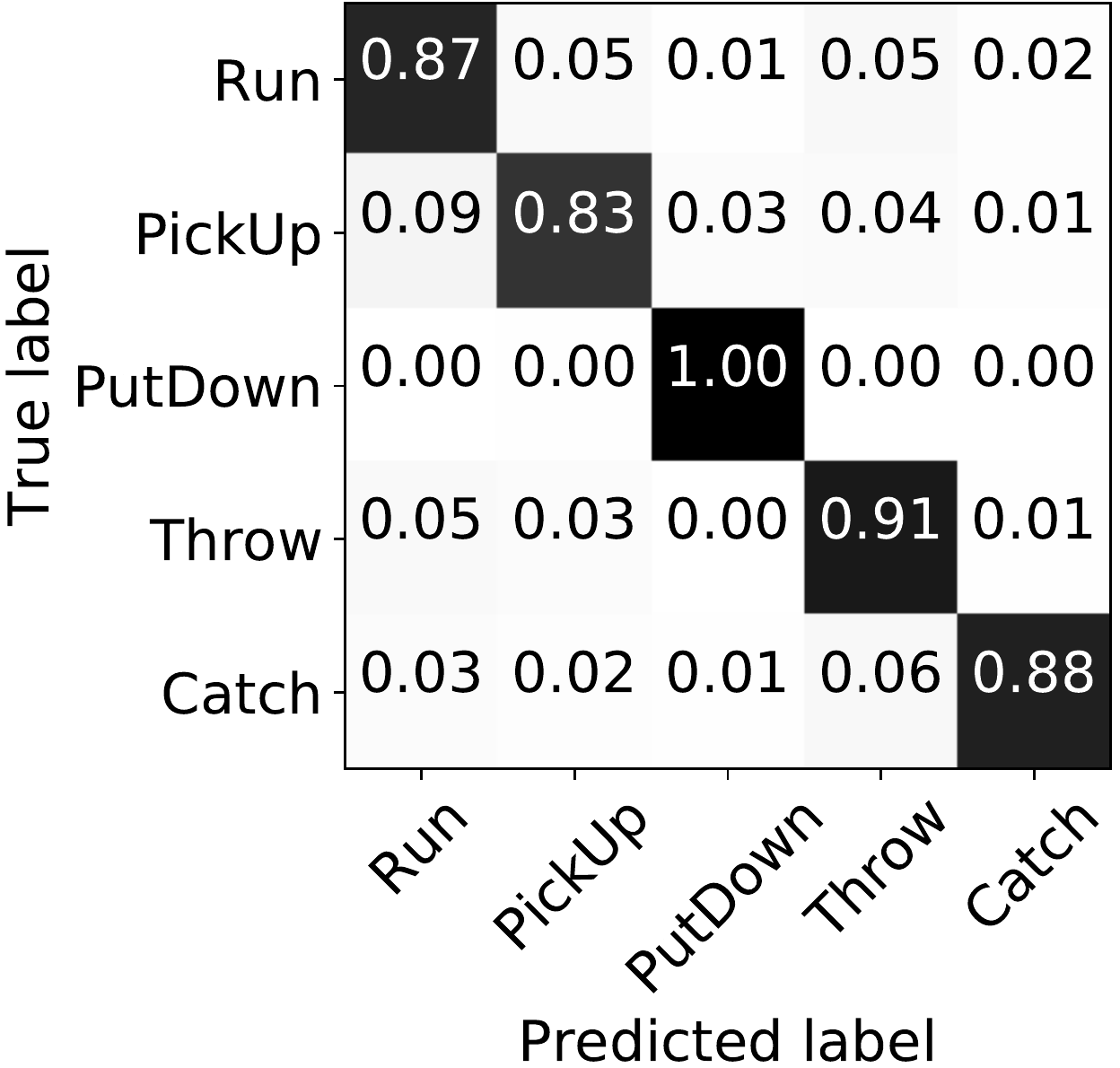}
\caption{Confusion matrices of action recognition on view-centric proposals (left) and scene-centric predictions (right).}
\label{fig:action-confusion-matrix}
\end{figure}

\begin{table*}[ptb]
\begin{center}
\setlength\tabcolsep{8pt}
\resizebox{\linewidth}{!}{
\begin{tabular}{l||l|cccccccccc}
\thickhline
\multirow{5}{*}{\textbf{CAMPUS}} & Methods & Gender & Long hair & Glasses & Hat & T-shirt &  Long sleeve & Shorts & Jeans & Long pants & mAP\\
\cline{2-12}\cline{2-12}
& view-centric &
	0.59 & 0.77 & 0.56 & 0.76 & 0.36 & 0.59 & 0.70 & 0.63 & 0.35 & 0.59 \\
& baseline-mean &   
	0.63 & 0.82 & 0.55 & 0.75 & 0.34 & 0.64 & 0.69 & 0.63 & 0.34 & 0.60 \\
& baseline-vote &   
	0.61 & 0.82 & 0.55 & 0.75 & 0.34 & 0.65 & 0.69 & 0.63 & 0.35 & 0.60 \\
& scene-centric &  
    0.76 & 0.82 & 0.62 & 0.80 & 0.40 & 0.62 & 0.76 & 0.62 & 0.24 & \best{0.63} \\
\thickhline
\multirow{5}{*}{\textbf{TUM Kitchen}} & Methods & Gender & Long hair & Glasses & Hat & T-shirt &  Long sleeve & Shorts & Jeans & Long pants & mAP\\
\cline{2-12}\cline{2-12}
& view-centric &
	 0.69 & 0.93 & 0.32 & 1.00 & 0.50 & 0.89 & 0.91 & 0.83 & 0.73 & 0.76 \\
& baseline-mean & 
     0.86 & 1.00 & 0.32 & 1.00 & 0.54 & 0.96 & 1.00 & 0.83 & 0.81 & 0.81 \\
& baseline-vote & 
	 0.64 & 1.00 & 0.32 & 1.00 & 0.32 & 0.93 & 1.00 & 0.83 & 0.76 & 0.76 \\
& scene-centric & 
	 0.96 & 0.98 & 0.32 & 1.00 &  0.77 & 0.96 & 0.94 & 0.83 & 0.83 & \best{0.84} \\
\thickhline
\end{tabular}
}
\end{center}
\caption{Quantitative comparisons of human attribute recognition on CAMPUS and TUM Kitchen datasets.}
\label{tab:attr}
\end{table*}

\begin{figure}[ptb]
\centering
\includegraphics[width=0.85\linewidth]{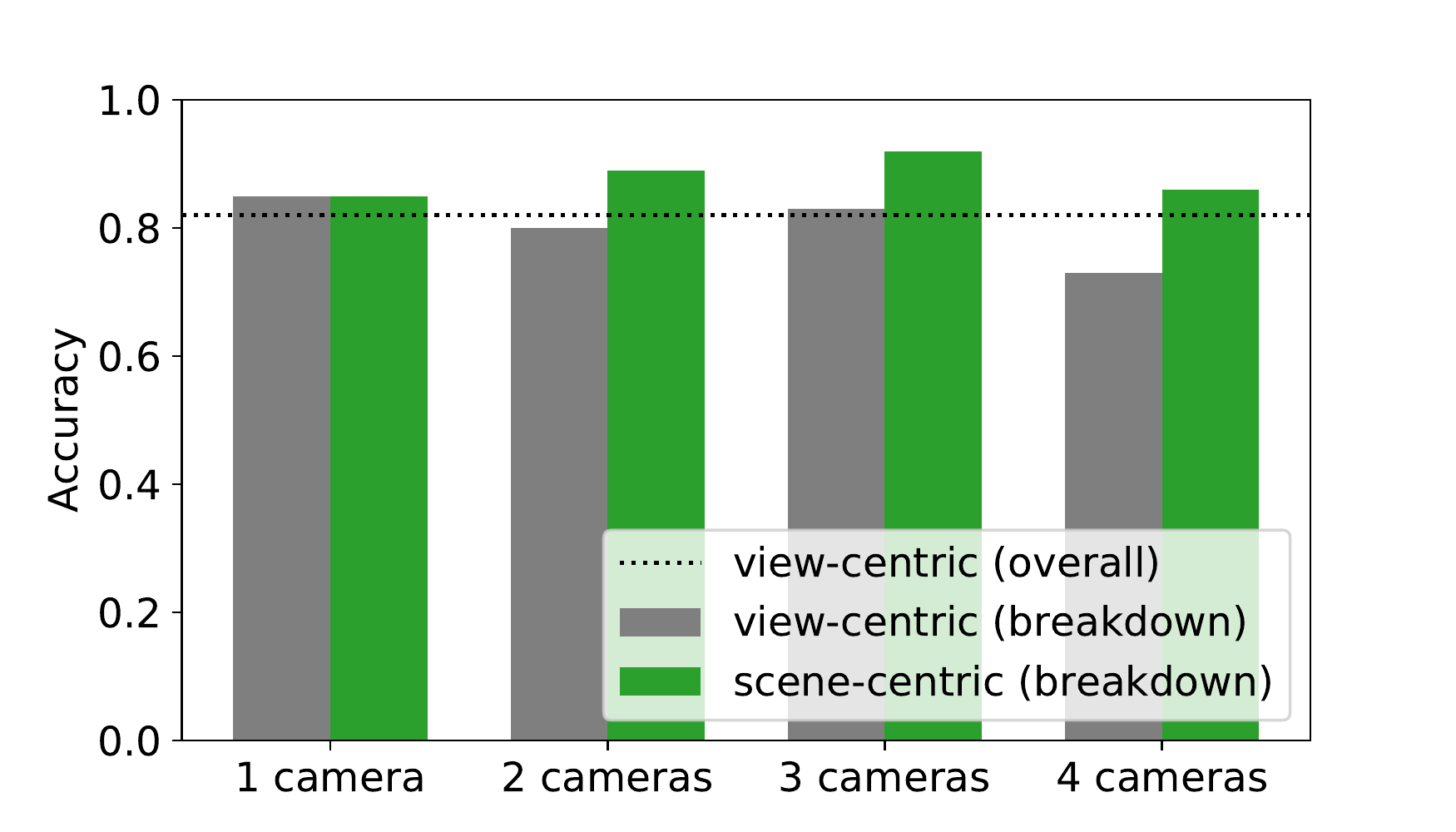}
\caption{The breakdown of action recognition accuracy according to the number of camera views in which each entity is observed.}
\label{fig:action-camera-breakdown}
\end{figure}

\begin{figure*}[ptb]
\centering
\includegraphics[width=\linewidth]{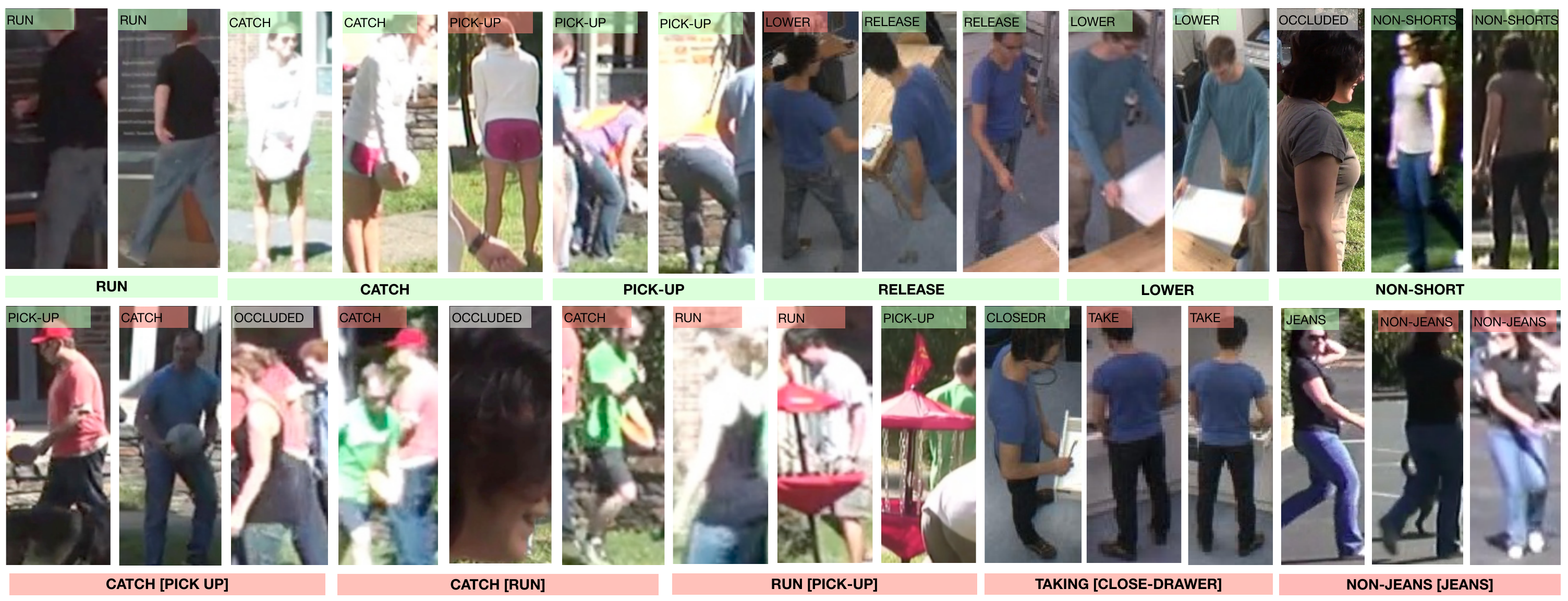}
\caption{Success (1st row) and failure examples (2nd row) of view-centric (labels overlaid on the images) and scene-centric predictions (labels beneath the images) of action and attribute recognition tasks. For failure examples, true labels are in the bracket. ``Occluded'' means that the locations of objects or parts are projected from scene locations and therefore no view-centric proposals are generated. Better viewed in color.}
\label{fig:examples}
\end{figure*}

\textbf{Action recognition}. View-centric action proposals are obtained from a fully-connected neural network with 5 hidden layers and 576 neurons which predicts action labels using human pose.
For the CAMPUS dataset, we collect additional annotations for 5 human action classes: Run, PickUp, PutDown, Throw, and Catch in total of 8,801 examples.
For the TUM Kitchen dataset, we evaluate on the 8 action categories: Reaching, TakingSomething, Lowering, Releasing, OpenDoor, CloseDoor, OpenDrawer, and CloseDrawer.
We measure both individual accuracies for each category as well as the overall accuracies across all categories. Table~\ref{tab:action} shows the performance of scene-centric predictions with view-centric proposals, and two additional fusing strategies as baselines. Concretely, the \textit{baseline-vote} strategy takes action predictions from multiple views and outputs the label with majority voting, while the \textit{baseline-mean} strategy assumes equal priors on all cameras and outputs the label with the highest averaged probability. When evaluating scene-centric predictions, we project scene-centric labels back to individual bounding boxes and calculate accuracies following the same procedure as evaluating view-centric proposals. Our joint parsing framework demonstrates improved results as it aggregates marginalized decisions made on individual views while also encourages solutions that comply with other tasks. Fig.~\ref{fig:action-confusion-matrix} compares the confusion matrix of view-centric proposals and scene-centric predictions after joint parsing for CAMPUS dataset.
To further understand the effect of multiple views, we break down classification accuracies by the number of cameras where persons are observed (Fig.~\ref{fig:action-camera-breakdown}). Observing an entity from more cameras generally leads to better performance, while too many conflicting observations may also cause degraded performance. Fig.~\ref{fig:examples} shows some success and failure examples.

\textbf{Human attribute recognition}. We follow the similar procedure as in the action recognition case above. Additional annotations for 9 different types of human attributes are collected for both CAMPUS and TUM Kitchen dataset. View-centric proposals and score are obtained from an attribute grammar model as in~\cite{park2016attr}. We measure performance with average precisions for each attribute categories as well as mean average precision (mAP) as in human attribute literatures. Scene-centric predictions are projected to bounding boxes in each views when calculating precisions. Table~\ref{tab:attr} shows quantitative comparisons between view-centric and scene-centric predictions. The same baseline fusing strategies as in the action recognition task are used. The scene-centric prediction outperforms the original proposals in  7 out of 9 categories while remains comparable in others. Notably, the CAMPUS dataset is harder than standard human attribute datasets because of occlusions, limited scales of humans, and irregular illumination conditions.

\subsection{Runtime}

With initial view-centric proposals precomputed, for a 3-minute scene shot by 4 cameras containing round 15 entities, our algorithm performs at 5 frames per second on average. With further optimization, our proposed method can run in real-time. Note that although the proposed framework uses a sampling-based method, using view-based proposals as initialization warm-starts the sampling procedure. Therefore, the overall runtime is significantly less than searching the entire solution space from scratch. For problems of a larger size, more efficient MCMC algorithms may be adopted. For example, the mini-batch acceptance testing technique~\cite{chen2016efficient} has demonstrated several order-of-magnitude speedups.

\section{Conclusion}

We represent a joint parsing framework that computes a hierarchy of parse graphs which represents a comprehensive understanding of cross-view videos. We explicitly specify various constraints that reflect the appearance and geometry correlations among objects across multiple views and the correlations among different semantic properties of objects. Experiments show that the joint parsing framework improves view-centric proposals and produces more accurate scene-centric predictions in various computer vision tasks.

We briefly discuss advantages of our joint parsing framework and potential future directions from two perspectives.

\subsubsection{Explicit Parsing} \label{sec:explicit}

While the end-to-end training paradigm is appealing in many \textit{data-rich} supervised learning scenarios, as an extension, leveraging loosely-coupled pre-trained modules and exploring commonsense constraints can be helpful when large-scale training data is not available or too expensive to collect in practice. For example, many applications in robotics and human-robot interaction domains share the same set of underlying perception units such as scene understanding, object recognition, etc. Training for every new scenarios entirely could end up with exponential number of possibilities. Leveraging pre-trained modules and explore correlation and constraints among them can be treated as a factorization of the problem space. Therefore, the explicit joint parsing scheme allows practitioners to leverage pre-trained modules and to build systems with an expanded skill set in a scalable manner.

\subsubsection{Interpretable Interface.} \label{sec:explainable}

Our joint parsing framework not only provides a comprehensive scene-centric understanding of the scene, moreover, the sence-centric spatio-temporal parse graph representation is an interpretable interface of computer vision models to users. In particular, we consider the following properties an explainable interface shall have apart from the correctness of answers:
\begin{itemize}
\item \textit{Relevance}: an agent shall recognize the intent of humans and provide information relevant to humans' questions and intents.
\item \textit{Self-explainability}: an agent shall provide information that can be interpreted by humans as how answers are derived. This criterion promotes humans' trust on an intelligent agent and enables sanity check on the answers.
\item \textit{Consistency}: answers provided by an agents shall be consistent throughout an interaction with humans and across multiple interaction sessions. Random or non-consistent behaviors cast doubts and confusions regarding the agent's functionality.
\item \textit{Capability}: an explainable interface shall help humans understand the boundary of capabilities of an agent and avoid blinded trusts.
\end{itemize}
Potential future directions include quantifying and evaluating the interpretability and user satisfaction by conducting user studies.


{
\small
\bibliographystyle{aaai}
\bibliography{scene_parsing}
}

\end{document}